\pdfoutput=1

\documentclass[11pt]{article}

\usepackage[]{acl}

\usepackage{times}
\usepackage{latexsym}

\usepackage[T1]{fontenc}

\usepackage[utf8]{inputenc}

\usepackage{microtype}

\usepackage{inconsolata}

\usepackage{framed}
\usepackage{mathtools}
\usepackage{graphics}
\usepackage{graphicx}
\usepackage{amsmath}
\usepackage{amsfonts,amssymb}
\usepackage{amssymb}
\usepackage{bbm}
\usepackage[ruled, linesnumbered]{algorithm2e}
\usepackage{amsmath}
\usepackage{booktabs}
\usepackage{multirow}
\usepackage{color}
\usepackage{xcolor}
\usepackage{pifont}
\usepackage{pgfplots}
\usepackage{makecell}
\usepackage{caption}
\usepackage{subcaption}
\usepackage{math_cmd}

\title{Sim-GPT: Text Similarity via GPT Annotated Data}

\author{
Shuhe Wang$^{\spadesuit}$,
Beiming Cao$^{\blacklozenge}$,
Shengyu Zhang$^{\clubsuit}$,
Xiaoya Li$^{\blacktriangledown}$\\
{\bf Jiwei Li$^{\clubsuit}$,
Fei Wu$^{\clubsuit}$,
Guoyin Wang$^{\bigstar}$,
Eduard Hovy$^{\blacktriangle}$}
}

\begin{document}
\maketitle
\begin{abstract}
Due to the lack of a large collection of high-quality labeled sentence pairs with textual similarity scores, existing approaches for Semantic Textual Similarity (STS) mostly rely on unsupervised techniques or training signals that are only partially correlated with textual similarity, e.g., NLI-based datasets.


To tackle this issue, 
in this paper, 
we propose 
the strategy of measuring text similarity via GPT annotated data
(Sim-GPT for short). 
The core idea of Sim-GPT is to generate data with STS labels using GPT-4, based on which an STS model is trained. 
Sim-GPT framework utilizes LLMs to provide a substantial amount of reliable annotated data filling the gap of the lack of training signals for STS.  
Sim-GPT is trained on 
a one-time generated dataset using BERT or RoBERTa as the backbone, which 
 offers long-term savings in cost and speed compared to repeatedly invoking LLMs for each sentence pair.

Trained on the examples from GPT-4 (371K), Sim-GPT yields SOTA performances on the widely-used seven STS benchmarks: +0.99 over supervised-SimCSE \cite{gao2021simcse}, and +0.42 over the current SOTA PromCSE \cite{jiang2022improved} model. To encourage further advancements of the field, we release both 
models and the 
371K annotated  examples from GPT-4. Code, models and annotated data are available at: \url{https://github.com/ShuheWang1998/Sim-GPT}.


\end{abstract}

\let\thefootnote\relax\footnotetext{$^{\spadesuit}$Peking University, $^{\clubsuit}$Zhejiang University, $^{\blacktriangledown}$Shannon.AI \\ $^{\blacklozenge}$University of Texas at Austin, $^{\blacktriangle}$University of Melbourne, \\ $^{\bigstar}$Bytedance}

\let\thefootnote\relax\footnotetext{wangshuhe@stu.pku.edu.cn,  xiaoya\_li@shannonai.com,\\ \{sy\_zhang, jiwei\_li, wufei\}@zju.edu.cn, beiming.cao @utexas.edu, guoyin.wang@bytedance.com, eduard.hovy @unimelb.edu.au}


\section{Introduction}
Measuring 
Semantic textual similarity (STS) \cite{devlin2018bert,liu2019roberta,reimers2019sentence,lan2019albert,gao2021simcse, liu2019roberta,jiang2022improved, seonwoo2022ranking, chuang2022diffcse,sun2022sentence, chen2023sub}
 between sentences
is an important task in natural language processing. 
A longstanding issue with STS is the lack of supervised training data: 
it is infeasible to obtain a large collection of 
high-quality
labeled sentence pairs with textual similarity scores.
Existing approaches either are in a unsupervised fashion \cite{jiang2019smart,wang2019structbert,lan2019albert,yang2019xlnet,raffel2020exploring,zaheer2020big} or rely on training signals that are partially correlated 
 with textual similarity\cite{reimers2019sentence,yan2021consert,gao2021simcse,jiang2022improved}, e..g, 
  SNLI\cite{bowman2015large} and MNLI\cite{williams2017broad} datasets whose annotations indicate whether the entailment or contradiction relationship holds between two sentences. 
Both unsupervised approaches and NLI-based approaches
suffer from the lack of \textsc{direct} supervised signals for training STS models. 

Large language models (LLMs) \cite{brown2020language,zhang2020unsupervised,rae2021scaling,chowdhery2022palm,tan2022exploring,ouyang2022training,openai2023gpt4,touvron2023llama} open a new door for 
the task of STS. 
With their impressive language comprehension abilities, these models have the ability to directly generate accurate semantic similarity scores for pairs of sentences. 
However, there are two main concerns when using LLMs to provide STS scores.
Firstly, it can be costly to use LLMs like GPT each time a new sentence pair is encountered. Secondly, the speed of LLMs is typically significantly slower compared to models like BERT\cite{devlin2018bert} or RoBERTa\cite{liu2019roberta}.

To better harness LLMs' ability for the STS task, 
in this paper, we propose Sim-GPT. 
Sim-GPT does not directly ask LLMs (e.g., GPT-4) to provide STS scores for a newly-encounter sentence pair. But rather,  it firstly asks LLMs to generate a relatively large set of training data; 
secondly, a smaller model (e.g., backboned by BERT or RoBERTa) is trained based on the synthesized data from LLMs;
At test time, the trained model is used for inference. 

Sim-GPT tackles the aforementioned issues simultaneously. It resolves the long-standing problem of the lack of training signals for STS by utilizing LLMs to provide a substantial amount of reliable supervision labels for training. Additionally, it addresses concerns regarding cost and speed. In terms of cost, while generating synthesized data using LLMs can still be expensive for large training sets, this cost is incurred only once. The trained model can then be repeatedly used by developers and the community without any additional expenses, unlike the scenario where the LLM needs to be invoked every time an STS score needs to be computed. As for speed, the inference process using a BERT or RoBERTa backboned model is significantly faster.

Trained on the 646K synthesized examples from GPT-4 (371K) and NLI (275K) data\footnote{SNLI\cite{bowman2015large} and MNLI\cite{williams2017broad}}, Sim-GPT yields SOTA performances on the widely-used seven STS benchmarks:
 +0.99 over supervised-SimCSE \cite{gao2021simcse}, and +0.42 over the current SOTA PromCSE \cite{jiang2022improved} model. 
The contribution of this work can be summarized as follows:
\begin{enumerate}
\item We propose Sim-GPT, a general framework that utilizes data produced by LLMs to train the STS model effectively.
\item We collect a dataset consisting of 371K annotated STS examples from GPT-4. This dataset will be made available to the community to encourage further advancements.
\item Sim-GPT achieves SOTA performances on the STS benchmark, demonstrating the effectiveness of the proposed framework. 
\end{enumerate}
The rest of this paper is organized as follows:
in Section \ref{related_work}, we go through related work. 
Section \ref{data_annotation} describes the data generation process in detail. 
Section \ref{contrastive_training} describes the model training process.
Section \ref{experiments} details experimental results.
Ablations studies are presented in Section \ref{ablation_study}, followed by a brief conclusion in Section \ref{conclusion}.

\begin{figure*}[htb]
\centering
    \includegraphics[scale=0.35]{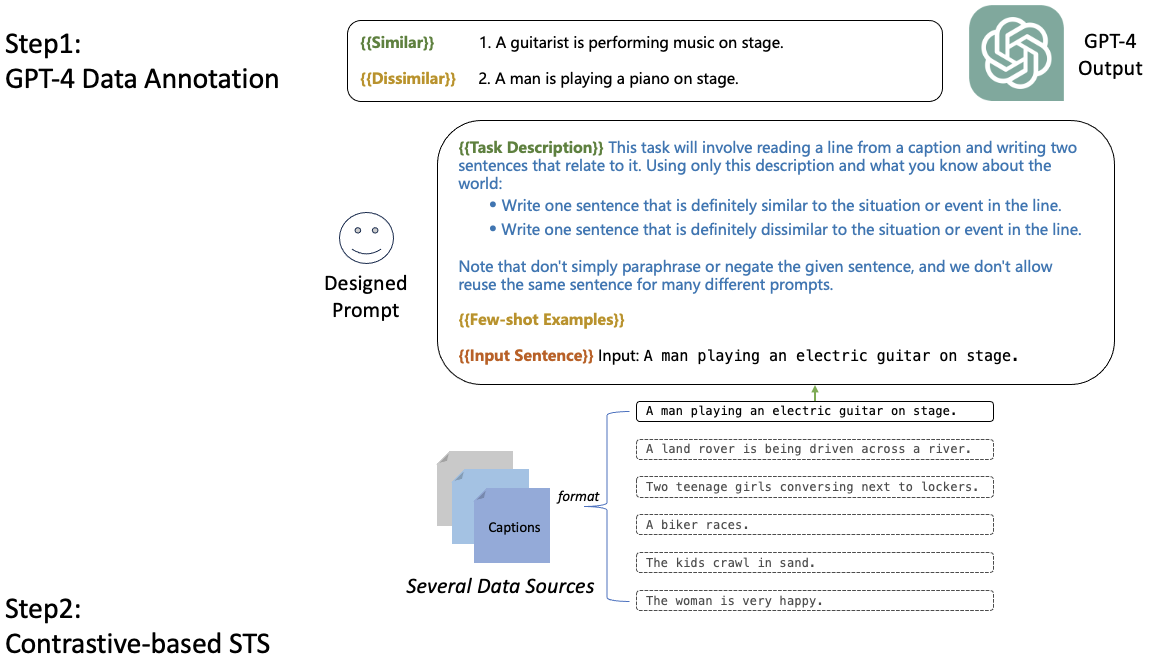}
    \includegraphics[scale=0.35]{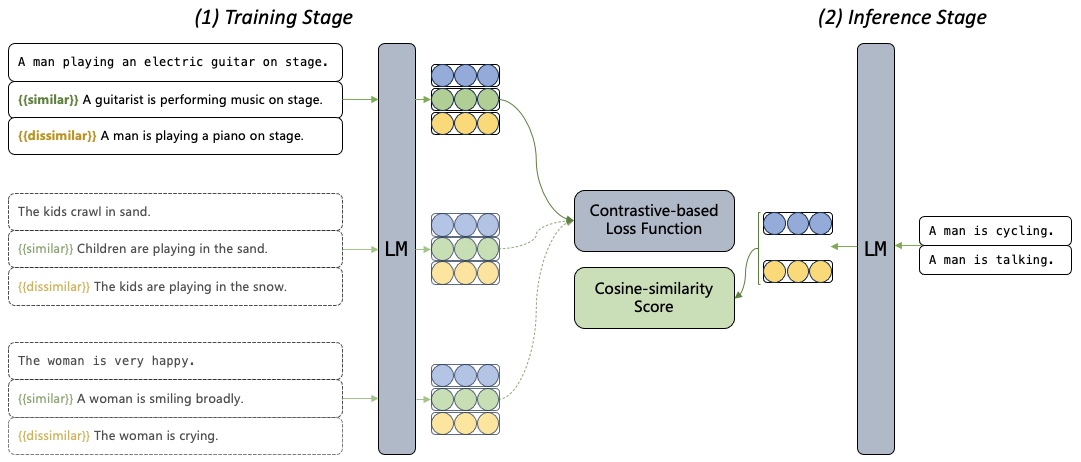}
  \caption{An example for the process of Sim-GPT, where the actual part represents the current input, e.g., the input sentence “\textit{A man playing an electric guitar on stage}” in the prompt, while the virtual part represents the input that are pending. \textbf{Step 1 GPT-4 Data Annotation}: Bottom-up, we instruct GPT-4 to annotate (\textit{similar, dissimilar}) format data for several input sentences. \textbf{Step 2 Contrastive-based STS}: The annotated data is used with similar sentence as positive and dissimilar sentence as negative to train STS models under the contrastive framework.}
  \label{fig:simcse_process}
\end{figure*}

\section{Related Work}
\label{related_work}
\subsection{Semantic Textual Similarity}
Due to the lack of supervised training data, previous approaches to STS relied on encoding sentence in high-dimensional spaces and enabling the computation of semantic similarity based on vector distances \cite{mikolov2013distributed,pennington2014glove,devlin2018bert,liu2019roberta}. Despite their strong performance, those unsupervised approaches are trained to capture general linguistic patterns in the language data lacking the guidance of specific STS information: similar granularity of two sentences.

To fill the gap of annotation data, \citet{bowman2015large} proposed the SNLI dataset and \citet{williams2017broad} proposed MNLI dataset covering a range of genres of spoken and written text. SNLI and MNLI consist of several hand-written sentence pairs, and each sentence pair is manually annotated with three labels according to whether there is a logical relationship between them: entailment, neutral and contradiction.

With the availability of NLI-data, many supervised approaches emerged promising to surpass their unsupervised counterparts. Fine-tuning Siamese networks \cite{neculoiu2016learning,conneau2017supervised,pontes2018predicting,ranasinghe2019semantic,benajiba2019siamese} that involve two parallel neural networks with shared weights, processing each sentence in a pair and learning sentence embeddings in a way that maximizes similarity for entailment sentences and minimizes it for contraditcion ones. Sentence-BERT \cite{reimers2019sentence} that derives from the BERT \cite{devlin2018bert} architecture, encoding semantically meaningful sentence embeddings and optimizing them with cosine-similarity. Contrastive-based models \cite{yan2021consert,kim2021self,gao2021simcse,jiang2022improved} that penalize the model for misclassifying similar sentence pairs as dissimilar, making similar sentences closer to each other in the embedding space, while dissimilar sentences are farther apart.

\subsection{Large Language Models}
LLMs are trained on vast volumes of text data and have emerged as a pivotal element in recent NLP studies. To improve the generation ability, some researchers explore to scale up LLMs, such as GPT-3 \citep{brown2020language} with 175B parameters, Gopher \citep{rae2021scaling} with 280B parameters and PaLM \citep{chowdhery2022palm} with 540B parameters. However, with the scaling up, diminishing margins are becoming more apparent. Thus, some researchers turned to explore how to efficiently utilize large training data and the learning ability of large parameters, including \citet{touvron2023llama} proposed LLaMA demonstrating that the performance of a 7B model continues to
improve even after 1T tokens and trained LLaMA-13B with only 13B parameters outperforming GPT-3 \citep{brown2020language}
(175B) on most benchmarks.
Others aims to train better LLMs with supervised learning or human instruction on each epoch \citep{wei2022finetuned,wang2022pre,mishra2022crosstask,ouyang2022training,zhang2023beam,zhang2023instruction}.

Given the impressive capability of LLMs to comprehend and produce human-like content, it's crucial to investigate their application to various downstream activities. One line of work aims to directly fine-tune LLMs on specific tasks, and to minimize fine-tune consumption, some of work focus on efficient fine-tuning such as partial parameters fine-tuning \citep{hu2021lora,li2021prefixtuning,liu2022fewshot,sun2023pushing,sun2023sentiment} and using memory mechanisms to cache past inputs \citep{wu2022memorizing,bulatov2022recurrent}. Rather than fine-tuning, there is a line of work aims to maximize the pre-trained knowledge of LLMs by prompts without altering the architecture \citep{brown2020language,wang2023gpt,sun2023text,wan2023gptre,qin2023chatgpt,li2023human}. Other lines includes knowledge distillation that use LLMs to teach smaller task-specific models \citep{hsieh2023distilling,dasgupta-etal-2023-cost,gu2023knowledge}, view LLMs as encoder embedding input words to vectors \citep{ge2023incontext} and using LLMs as auxiliary parts to activate the ability of supervised models \citep{sun2023test}.

\renewcommand{\dblfloatpagefraction}{.9}
\begin{figure*}
	\begin{framed}
	\footnotesize
		\underline{\textbf{Captions:}} \\
      	\text{A statue at a museum that no seems to be looking at.} \\
      	\text{A man reads the paper in a bar with green lighting.} \\
      	\text{A woman wearing a ball cap squats down to touch the cracked earth.} \\
      	\text{Two children re laying on a rug with some wooden bricks laid out in a square between them.} \\\\
      	\underline{\textbf{Questions:}} \\
      	\text{What is the step by step guide to invest in share market in india?} \\
      	\text{How can I increase the speed of my internet connection while using a VPN?} \\
      	\text{Motorola (company): Can I hack my Charter Motorolla DCX3400?} \\
      	\text{Method to find separation of slits using fresnel biprism?} \\\\
      	\underline{\textbf{Multi-genre Long Sentences:}} \\
      	\text{Networks also undergo a discrete training phase, as opposed to engaging in ongoing learning.} \\ 
      	\text{Take advantage of the current price and include it in your purchase if it catches your interest.} \\
      	\text{Vietnam on the verge of securing its position as the second largest economy in Southeast Asia...} \\
      	\text{We would appreciate it if you could share your knowledge about the music scene in Dallas, TX.} \\
    \end{framed}
\caption{An example of our collected data, which consists of three types of sources: captions, questions and multi-genre long sentences.}
\label{fig:data_sourc_example}
\end{figure*}

\section{Sim-GPT: GPT-4 Data Annotation}
\label{data_annotation}
To address the longstanding issue of the STS: the lack of supervised training data, we propose Sim-GPT, which can be decomposed into two steps: (1) GPT-4 Data Annotation, utilize LLMs to generate a relatively large set of training data; and (2) Contrastive-based STS, utilize the annotated data to train smaller models (backboned by contrastive learning), and then at test time do inference on the trained model. Illustrations are shown in Figure \ref{fig:2_few_shot_example}.

In this section, we describe the step “(1) GPT-4 Data Annotation”, and for step “(2) Contrastive-based STS” we put it in Section \ref{contrastive_training}.

\subsection{Overview}
Formally, GPT-4 is anticipated to generate (\textit{similar sentence}, \textit{dissimilar sentence}) format pair for each given sentence. Thus, below, we sequentially focus on collecting sentences from multi-genre sources and prompt design in guiding GPT-4 annotation.

\begin{table*}[th!]
    \centering
    \resizebox{1.0\textwidth}{!}{
    \begin{tabular}{l|ccccc}\toprule
        \textbf{Data} & \textbf{Source/Annotation} & \textbf{Sentence Number} & \textbf{Max Length} & \textbf{Min Length} & \textbf{Mean Tokens} \\\midrule
        \multirow{4}{*}{\bf Captions} & \textit{source} & 154933 & 82 & 2 & 13.5 \\\cline{2-6}
        & \textit{origin} & 138100 & 82 & 2 & 13.5 \\
        & \textit{similar} & 138100 & 27 & 4 & 10.4 \\
        & \textit{dissimilar} & 138100 & 28 & 4 & 9.3 \\\midrule
        
        \multirow{4}{*}{\bf Questions} & \textit{source} & 100000 & 87 & 2 & 11.4 \\\cline{2-6}
        & \textit{origin} & 96751 & 87 & 2 & 11.4 \\
        & \textit{similar} & 96751 & 65 & 2 & 12.2 \\
        & \textit{dissimilar} & 96751 & 50 & 2 & 10.4 \\\midrule
        
        \multirow{4}{*}{\bf Multi-genre Sentences} & \textit{source} & 150000 & 182 & 1 & 11.5 \\\cline{2-6}
        & \textit{origin} & 136545 & 182 & 1 & 11.6 \\
        & \textit{similar} & 136545 & 56 & 2 & 10.5 \\
        & \textit{dissimilar} & 136545 & 53 & 2 & 10.2 \\\bottomrule
    \end{tabular}
    }
    \caption{Key statistics for the raw sentence pairs in our collected data. For the annotated data, we report the triplet (origin, similar, dissimilar) separately. Noted that to ensure the resulting models are not overly specialized in understanding or responding to questions alone, we randomly selected 100K questions from the total pool of 400K for labeling.}
    \label{information_for_sources}
\end{table*}

\subsection{Data Sources}
In order to ensure that the training data generated by GPT-4 is comprehensive, we desire a wide range of input sentences.
Input sentences come from three sources, the details of which are shown in Figure \ref{fig:data_sourc_example}.

%

(1) \textbf{Captions}, which refer to hand-written texts describing related images or videos.
We choose the Flickr30k \cite{young2014image} corpus as the caption source. It consists of 31,783 images collected from everyday scenes containing people engaged in various activities, and each image in the Flickr30k dataset is accompanied by five different textual captions, providing a rich set of linguistic data to complement the visual information. 

(2) \textbf{Questions}, which refer to hand-written questions about specific information or insights.
We choose the Quora Question Pairs (QQP) \cite{quora-question-pairs} dataset as the hand-written question source. It is a collection of 400K user question-answers from Quora website involving a wide range of topics, e.g, education (“\textit{Is it a good idea to go to medical school without a Science degree?}”), math (“\textit{If 3x + 9y = 7x + y. then 8y=?}”) and nationality (“\textit{Why do Slavs squat?}”).

(3) \textbf{Multi-genre Long Sentences}, which ranges over a variety of spoken and written sentences.
We use a subset of redpajamas\footnote{\url{https://github.com/togethercomputer/RedPajama-Data}} which contains 150K long sentences from our real-world applications as multi-genre sources.

Table \ref{information_for_sources} shows some key statistics about the collected data. It's worth noting that the amount of annotated data is less than the amount of source data. The main reason lies that GPT-4 has the probability to violate the few-shot format outputting “\textit{weird}” results, thus, we write scripts to remove those unqualified GPT-4 outputs.

Here, we have collected a large amount of sentences from three types of sources. Next, we will describe the designed prompt to guide GPT-4 to generate decent annotation data.


\renewcommand{\dblfloatpagefraction}{.9}
\begin{figure*}
	\begin{framed}
	\footnotesize
This task will involve reading a line from a caption and writing two sentences that relate to it. Using only this description and what you know about the world: \\

Write one sentence that is definitely similar to the situation or event in the line. \\
Write one sentence that is definitely dissimilar to the situation or event in the line. \\

Note that don't simply paraphrase or negate the given sentence, and we don't allow reuse the same sentence for many different prompts. \\

      	\textit{Input: }\text{A land rover is being driven across a river.} \\
      	\textit{Output: } \\
      	\text{1. A vehicle is crossing a river.} \\
      	\text{2. A sedan is stuck in the middle of a river.} \\\\
      	\textit{Input: }\text{Two teenage girls conversing next to lockers.} \\
      	\textit{Output: } \\
      	\text{1. People talking next to lockers.} \\
      	\text{2. Girls talking next to the toilet.} \\
      	
      	\textit{Input:}
    \end{framed}
\caption{The 2-shot example of the designed prompt for our caption source data.}
\label{fig:2_few_shot_example}
\end{figure*}

\subsection{Prompt Design}
Given the input sentence, we instruct GPT-4 to generate two sentences: one is a similar sentence that semantically follows the input sentence; the other is a dissimilar sentence in semantics, e.g., 

\textit{Input Sentence}: Dogs run outdoors.

\textit{Similar Sentence}: Dogs are running.

\textit{Dissimilar Sentence}: There aren't any animals.

Shown in Figure \ref{fig:simcse_process}, the designed prompt consists of three parts: (1) Task Description; (2) Few-shot Examples; and (3) Input Sentence:

\subsubsection{Task Description}
Task description in the prompts for different sources are a little bit different. We use the caption for illustration, which is designed as follows:

\begin{framed}
\noindent This task will involve reading a line from a caption and writing two sentences that relate to it. Using only this description and what you know about the world:
	\begin{itemize}
		\item Write one sentence that is definitely similar to the situation or event in the line.
		\item Write one sentence that is definitely dissimilar to the situation or event in the line.
	\end{itemize}
Note that don't simply paraphrase or negate the given sentence, and we don't allow reuse the same sentence for many different prompts.
\end{framed}

This is the version for caption sources. For the version of question sources and multi-genre sources, they are akin to the caption version and are put in Appendix \ref{appendix_prompt}.

The core of the task description lies in three aspects:

(1) The first aspect,

“\textit{This task will involve reading a line from a caption and writing two sentences that relate to it. Using only this description and what you know about the world.}”

\noindent emphasizes that the following task is a STS task, and the annotation shouldn't be influenced or biased by the original information contained in the photograph (e.g., place names or author names about the photograph).

(2) The second aspect,

“\textit{Write one sentence that is definitely [similar / dissimilar] to the situation or event in the line.}”

\noindent indicates that GPT-4 should generate two sentences: One describes the same object as the input caption (as the similar sentence), and the other describes partially or totally different object as the input caption (as the dissimilar sentence).

(3) The third aspect,

“\textit{Note that don't simply paraphrase or negate the given sentence, and we don't allow reuse the same sentence for many different prompts.}”

\noindent avoids that GPT-4 just alter the syntactic structure of the input sentence, for example:

\textit{Input}: A man is in front of the tower.

\textit{Output}:

1. The tower is in front of a man.

2. A man isn't in front of the tower.

\subsubsection{Few-shot Examples}
Few-shot examples mainly serve as two purposes: (1) it provides a direct interpretation about the task description as the complexity of the above description; (2) it regulates the format of the GPT-4 outputs as the feature that GPT-4 mimics the format of the few-shot. 

For the format, we sequentially pack $k$ examples in the few-shot, and each example consists of one sentence as the input and two sentences (\textit{similar}/\textit{dissimilar}) as the output:

\begin{equation}
\nonumber
    \begin{aligned}
      	\textit{Input: }&\text{[Example Sentence]}_{1} \\
      	\textit{Output: }&\text{1. [Similar Sentence]}_{1} \\
      	&\text{2. [Dissimilar Sentence]}_{1} \\
      	&\cdots \\
      	\textit{Input: }&\text{[Example Sentence]}_{k} \\
      	\textit{Output: }&\text{1. [Similar Sentence]}_{k} \\
      	&\text{2. [Dissimilar Sentence]}_{k}
    \end{aligned}
\end{equation}

For each example, we manually select sentence in our data sources and write related similar/dissimilar sentences. Here, we display the designed prompt with 2-shot for our caption source data in Figure \ref{fig:2_few_shot_example}, and put the full designed prompt used for other data sources in Appendix \ref{appendix_prompt}.

\subsubsection{Input Sentence}
Input sentence feeds the current input sentence into the GPT-4 expecting it to output sequence according to the above defined format:

\begin{equation}
\nonumber
    \begin{aligned}
      	&\textit{Input: }\text{[Current Input]} \\
      	&\textit{Output: }
    \end{aligned}
\end{equation}
where \textit{Output} is the signal for the GPT-4 to output annotated results.

Here, we have completed instructions for GPT-4 annotating (\textit{similar}, \textit{dissimilar}) format data from various sources, and below we continue to describe the application of those annotated data in contrastive-based STS models.

\section{Sim-GPT: Contrastive-based STS Learning}
\label{contrastive_training}
Here, we describe the model training and inference process based on the annotated data.
In this work, we use the SimCSE framework \cite{gao2021simcse} as the backbone, which is a contrastive-based framework aiming to effectively pull semantically close similar sentences together and push apart dissimilar sentences. 

\subsection{SimCSE} 
Formally, supposed that $\mathcal{D}=\{(x_{i},x_{i}^{+},x_{i}^{-})\}_{i=1}^{m}$ represents the training set, where each sentence $x_{i}$ is packed with one similar sentence $x_{i}^{+}$ (semantically close) and one dissimilar sentence $x_{i}^{-}$ (semantically opposite), and $m$ denotes the size of the training set. 
The similar sentence and the dissimilar sentence can be directly taken from the generated training dataset. 
We first encode each triplet $(x,x^{+},x^{-})$ using the RoBERTa\cite{liu2019roberta} and generating vector representation triplet $(h,h^{+},h^{-})$. Then, the pre-trained RoBERTa is fine-tuned to optimize the contrastive-based loss function as:

\begin{equation}
\begin{aligned}
	&\mathcal{L}_{CL} = \\
	&-\log\frac{e^{sim(h_{i},h_{i}^{+})/\tau}}{\sum_{j=1}^{N}(e^{sim(h_{j},h_{j}^{+})/\tau} + \alpha*e^{sim(h_{i},h_{j}^{-})/\tau})}
\end{aligned}
\end{equation}
where $sim(h,h^{+})/\tau$ denotes the cosine similarity between sentence $x$ and $x^{+}$ with $\tau$ as the normalization parameter, and $\alpha$ denotes the weight of the dissimilar sentence.

During the inference stage, given the test input sentences pair $(x,y)$, we first send them into the fine-tuned RoBERTa generating the vector representations pair $(h_{x},h_{y})$. Then, the cosine similarity score is computed between $h_{x}$ and $h_{y}$ as the final similarity score of the sentences pair $(x,y)$.

Furthermore, in our evaluations of STS tasks where each sentence pair is assigned a score from 0 to 5, we will adjust/normalize the produced cosine similarity scores to fit within the 0 to 5 range.

\subsection{PromCSE} 
A follow-up work of SimCSE, which is called PromCSE \cite{jiang2019smart}, addresses the 
bottleneck of SimCSE because of the insufficient ability of the contrastive-based loss function $\mathcal{L}_{CL}$ in separating hard negatives and positives by expanding $\mathcal{L}_{CL}$ to:
 
\begin{equation}
\begin{aligned}
	&\mathcal{L} = \mathcal{L}_{CL} + \lambda \cdot \mathcal{L}_{M}, \\
	&\mathcal{L}_{M} = \max (0, m + sim(h_{i},h_{i}^{*}) - sim(h_{i},h_{i}^{+})), \\
	&h_{i}^{*} = \mathop{\arg\min}_{1\leq j\leq N, j\ne i} sim(h_{i},h_{j})
\end{aligned}
\end{equation}
where $sim$ denotes the cosine similarity, and $m$, $\lambda$ are two hyperparameters to amplify the differences between negatives and positives.

PromCSE achieves the current SOTA performance on supervised STS tasks, thus, 
in this work,
we also use PromCSE as the backbone, trained on the generated data from GPT-4. 
For the training and inference stage, PromCSE aligns with the above SimCSE, which is omitted for brevity.

\section{Experiments}
\label{experiments}
\subsection{Evaluation Settings}
To evaluate the effectiveness of Sim-GPT, we train the supervised-SimCSE and the current SOTA PromCSE model on the annotated Sim-GPT data, and conduct evaluations on 7 STS tasks. Parameters and settings are detailed below.

\paragraph{GPT Parameters.} We use GPT-4 \cite{openai2023gpt4} to generate annotated data. For the accessing parameters, we set: 

\textit{model = gpt-4}, \textit{temperature = }0, \textit{top\_p = }1, \textit{frequency\_penalty = }0, \textit{presence\_penalty = }0, \textit{max\_token = }4096

\noindent For the part of few-shot, we manually write 8 few-shot examples for each type of data source, which can be found in Appendix \ref{appendix_prompt}.

\paragraph{Training Details.} We follow default settings in SimCSE \cite{gao2021simcse} and PromCSE \cite{jiang2019smart} using the pre-trained RoBERTa (cased) \cite{liu2019roberta} and taking the $[CLS]$ representation as the sentence embedding. For training parameters, we adjust to the best as:

\begin{table}[ht]
    \centering
    \resizebox{.48\textwidth}{!}{
    \begin{tabular}{lcccc}\toprule
        & \multicolumn{2}{c}{{\bf SimCSE}} & \multicolumn{2}{c}{{\bf PromCSE}} \\
        & base & large & base & large \\\midrule
        Batch size & 512 & 512 & 512 & 256 \\\midrule
        Learning rate & 5e-5 & 1e-5 & 1e-2 & 5e-3 \\\midrule
        Negative weight ($\alpha$) & 1 & 0.8 & 0.8 & 0.8 \\\bottomrule
    \end{tabular}
    }
\end{table}

\begin{table*}[th!]
    \centering
    \resizebox{1.\textwidth}{!}{
    \begin{tabular}{lcccccccc}\toprule
        \textbf{Model} & \textbf{STS12} & \textbf{STS13} & \textbf{STS14} & \textbf{STS15} & \textbf{STS16} & \textbf{STS-B} & \textbf{SICK-R} & \textbf{Avg} \\\midrule
        \multicolumn{9}{c}{\it In-Context Learning with GPT-4} \\\midrule
        8-shot & 61.18 & 83.09 & 71.04 & 69.85 & 74.29 & 80.61 & 67.73 & 72.54 \\
        16-shot & 66.33 & 78.04 & 73.20 & 80.02 & 76.07 & 73.25 & 69.90 & 73.83 \\
        32-shot & 63.03 & 78.94 & 73.79 & 77.07 & 76.80 & 82.06 & 69.47 & 74.45 \\\midrule
        \multicolumn{9}{c}{\it Unsupervised Models} \\\midrule
        Word Embeddings (GloVe) $^{\spadesuit}$ & 55.14 & 70.66 & 59.73 & 68.25 & 63.66 & 58.02 & 53.76 & 61.32 \\
        BERT$_{\text{base}}^{\bigstar}$ (average word embeddings) & 39.70 & 59.38 & 49.67 & 66.03 & 66.19 & 53.87 & 62.06 & 56.70 \\
        $\text{CT-BERT}_{\text{base}}^{\bigstar}$ & 61.63 & 76.80 & 68.47 & 77.50 & 76.48 & 74.31 & 69.19 & 72.05 \\
        PromCSE-$\text{BERT}_{\text{base}}^{\blacktriangle}$ & 73.03 & 85.18 & 76.70 & 84.19 & 79.69 & 80.62 & 70.00 & 78.49 \\
        SimCSE-$\text{RoBERTa}_{\text{large}}^{\bigstar}$ & 72.86 & 83.99 & 75.62 & 84.77 & 81.80 & 81.98 & 71.26 & 78.90 \\\midrule
        \multicolumn{9}{c}{\it Supervised Models} \\\midrule
        InferSent-GloVe$^{\spadesuit}$ & 52.86 & 66.75 & 62.15 & 72.77 & 66.87 & 68.03 & 65.65 & 65.01 \\
        Universal Sentence Encoder$^{\spadesuit}$ & 64.49 & 67.80 & 64.61 & 76.83 & 73.18 & 74.92 & 76.69 & 71.22 \\
        $\text{SRoBERTa}_{\text{base}}^{\spadesuit}$ & 71.54 & 72.49 & 70.80 & 78.74 & 73.69 &  77.77 & 74.46 & 74.21 \\
        $\text{SBERT}_{\text{base}}^{\spadesuit}$ & 70.97 & 76.53 & 73.19 &  79.09 & 74.30 & 77.03 &  72.91 & 74.89 \\
        $\text{CT-SBERT}_{\text{base}}^{\bigstar}$ & 74.84 & 83.20 & 78.07 & 83.84 & 77.93 & 81.46 & 76.42 & 79.39 \\\midrule
        \multicolumn{9}{c}{\it Supervised Models based on Our Sim-GPT Framework} \\\midrule
       	SimCSE-$\text{RoBERTa}_{\text{base}}^{\bigstar}$ & 76.53 & 85.21 & \textbf{80.95} & 86.03 & 82.57 & 85.83 & 80.50 & 82.52 \\
       	* Sim-GPT + SimCSE-$\text{RoBERTa}_{\text{base}}$ & \textbf{77.65 (+1.12)} & \textbf{86.15 (+0.94)} & 80.58 (-0.37) & \textbf{86.47 (+0.44)} & \textbf{84.08 (+1.51)} & \textbf{86.20 (+0.37)} & \textbf{80.88 (+0.38)} & \textbf{83.14 (+0.62)} \\
       	SimCSE-$\text{RoBERTa}_{\text{large}}^{\bigstar}$ & 77.46 & 87.27 & 82.36 & 86.66 & 83.93 & 86.70 & \textbf{81.95} & 83.76 \\
       	* Sim-GPT + SimCSE-$\text{RoBERTa}_{\text{large}}$ & \textbf{78.79 (+1.33)} & \textbf{88.22 (+0.95)} & \textbf{83.48 (+1.12)} & \textbf{88.32 (+1.66)} & \textbf{85.48 (+1.55)} & \textbf{87.91 (+1.21)} & 81.07 (-0.88) & \textbf{84.75 (+0.99)} \\\midrule
		PromCSE-$\text{RoBERTa}_{\text{base}}^{\blacktriangle}$ & 77.51 & 86.15 & \textbf{81.59} & 86.92 & 83.81 & 86.35 & \textbf{80.49} & 83.26 \\
		* Sim-GPT + PromCSE-$\text{RoBERTa}_{\text{base}}$ & \textbf{77.74 (+0.23)} & \textbf{86.82 (+0.77)} & 81.36 (-0.23) & \textbf{87.01 (+0.09)} & \textbf{84.58 (+0.77)} & \textbf{86.98 (+0.63)} & 80.48 (-0.01) & \textbf{83.57 (+0.31)} \\
		PromCSE-$\text{RoBERTa}_{\text{large}}^{\blacktriangle}$ & 79.56 & \textbf{88.97} & 83.81 & 88.08 & 84.96 & 87.87 & 82.43 & 85.10 \\
       	* Sim-GPT + PromCSE-$\text{RoBERTa}_{\text{large}}$ & \textbf{79.92 (+0.36)} & 88.87 (-0.10) & \textbf{84.29 (+0.48)} & \textbf{88.64 (+0.56)} & \textbf{85.94 (+0.98)} & \textbf{88.18 (+0.31)} & \textbf{82.79 (+0.36)} & \textbf{85.52 (+0.42)} \\\bottomrule
      \end{tabular}
    }
    \caption{Results on STS tasks, and we highlight the highest numbers among models with the same framework. $\spadesuit$: results from \citet{reimers2019sentence}, results from \citet{gao2021simcse}, and $\blacktriangle$: results from \citet{jiang2022improved}.}
    \label{main_results}
\end{table*}

\paragraph{Evaluation Tasks.} Following the previous work \cite{reimers2019sentence,gao2021simcse,chuang2022diffcse,jiang2022improved}, we conduct experiments on 7 STS tasks: STS12-16 \cite{agirre2012semeval,agirre2013sem,agirre2014semeval,agirre2015semeval,agirre2016semeval} , STS Benchmark \cite{cer2017semeval} and SICK-Relatedness \cite{marelli-etal-2014-sick}. 

These tasks consist of a large amount of sentence pairs, each of which requires the evaluated model to produce a similarity score ranging from 0 to 5. For the metric, the Spearman’s correlation is reported for all experiments.

\paragraph{Baselines.}
We compare with both unsupervised and supervised models: 

For unsupervised models, we choose two methods based on word embeddings: GloVe embeddings \cite{pennington2014glove} and BERT embeddings \cite{devlin2018bert}; and three methods based on contrastive learning: CT-BERT \cite{carlsson2020semantic}, SimCSE \cite{gao2021simcse} and PromCSE \cite{jiang2022improved}. 

For supervised models, we choose three methods based on siamese neural networks: InferSent-GloVe \cite{conneau2017supervised}, Universal Sentence Encoder \cite{cer2018universal} and SBERT/SRoBERTa \cite{reimers2019sentence}; and two methods based on contrastive learning: CT-SBERT \cite{carlsson2020semantic}, SimCSE \cite{gao2021simcse} and the current supervised SOTA model PromCSE \cite{jiang2022improved}. Only the InferSent-GloVe model is trained on the SNLI \cite{bowman2015large} dataset, all the others are trained on SNLI+MNLI \cite{williams2017broad} dataset.

Additionally, we conduct experiments to contrast Sim-GPT against leveraging GPT-4 to directly generate semantic similarity scores for sentence pairs based on the given prompt. For more details about the accessing prompt, please check Appendix \ref{fig:context_learning_8}.

\subsection{Main Results}
Table \ref{main_results} shows evaluation results on 7 STS tasks, where we can observe: 
 
(1) The average score achieved by the in-context learning-based method is significantly lower compared to that of our Sim-GPT approach, e.g., with 32-shot, GPT-4 achieves an average score of 74.45, while the $\text{SimCSE-RoBERTa}_{\text{large}}$ model trained on our Sim-GPT annotated data achieves an average score of 84.75 (+10.3). This is because the number of illustration examples in in-context learning is limited, while the proposed Sim-GPT is designed to take the advantage of large amount of training examples.

(2) Focusing on a specific task, as the few-shot changes, the score achieved by the in-context learning-based method fluctuates dramatically. 
The dramatic fluctuations in scores suggest that in-context learning based STS approaches are highly sensitive to different few-shot selections, and are not stable over different few-shot settings. In contrast, in Section \ref{ablation_study_few_shot}, we do the ablation study by changing few-shot examples of Sim-GPT, showing that our Sim-GPT is highly stable in performance across different few-shot selections.

(3) There is a significant improvement by training supervised models on the Sim-GPT annotated data: 83.76 v.s. 84.75 (+0.99) based on the $\text{SimCSE-RoBERTa}_{\text{large}}$ model, and 85.10 v.s. 85.52 (+0.42) based on the $\text{PromCSE-RoBERTa}_{\text{large}}$ model. These results not only illustrate the effectiveness of Sim-GPT in annotating STS data but also demonstrate the flexibility of Sim-GPT in the selection of backboned models.

(4) Sim-GPT achieves the best average score of 85.52, which +0.42 over the current SOTA supervised PromCSE \cite{jiang2019smart} model. This SOTA performance further illustrates the effectiveness of the proposed Sim-GPT framework, encouraging future advancements based on the same idea as Sim-GPT.

\section{Ablation Study}
\label{ablation_study}
We do ablation studies on: (1) using different data sources; and (2) changing few-shot examples. All results are reported based on STS Benchmark development set using default GPT-4 settings and $\text{SimCSE-RoBERTa}_{\it base}$ parameters.

\subsection{Data Sources Permutation}
It's crucial to visually compare the contribution of each data source to the STS task.
Thus, we 
use different combinations of the
 three data sources: (1) Captions, (2) Questions, (3) Multi-genre Sentences, (4) Captions+Questions, (5) Captions+Multi-genre Sentences, (6) Questions+Multi-genre Sentences, and (7) Captions+Questions+Multi-genre Sentences. 

Table \ref{abalation_study_data_sources} shows results for different combinations on the STS Benchmark development set, and we can clearly observe that  more data  leads to  higher scores, e.g., combination “Captions + Questions + Multi-genre Sentences” leads to the score of 85.9, which is higher than the score 83.88 for “Captions + Questions” and the score 83.54 for “Captions” alone.

\begin{table}[th!]
    \tiny
    \centering
    \resizebox{.49\textwidth}{!}{
    \begin{tabular}{lc}\toprule
        \textbf{Combinations} & \textbf{Results} \\\midrule
        Captions & 83.54 \\
        Questions & 81.38 \\
        Multi-genre Sentences & 83.81 \\\midrule
        Captions + Questions & 83.88 \\
        Captions + Multi-genre Sentences & 84.96 \\
        Questions + Multi-genre Sentences & 84.02 \\\midrule
        Captions + Questions & \multirow{2}{*}{\bf 85.9} \\
        + Multi-genre Sentences &  \\\bottomrule
    \end{tabular}
    }
    \caption{STS Benchmark development results for different combinations of data sources.}
    \label{abalation_study_data_sources}
\end{table}

\begin{table}[th!]
    \tiny
    \centering
    \resizebox{.49\textwidth}{!}{
    \begin{tabular}{lc}\toprule
        \textbf{Few-shot Variation} & \textbf{STS-B Results} \\\midrule
        \text{Sim-GPT} & \textbf{83.54} \\\midrule
        \text{Changing few-shot number} & 83.41 \\
        \text{Changing few-shot examples} & 83.44 \\\bottomrule
    \end{tabular}
    }
    \caption{STS Benchmark development results with different types of few-shot examples. \textbf{Sim-GPT}: the used caption 8-shot examples in Sim-GPT.\textbf{Changing few-shot number}: we manually expand 8-shot examples of Sim-GPT to 16-shot. \textbf{Changing few-shot examples}: we alter caption 8-shot examples in Sim-GPT with multi-genre 8-shot examples.}
    \label{ablation_few_shot_type}
\end{table}

\subsection{Few-shot Variation in Data Generation}
\label{ablation_study_few_shot}
Commonly, the output of LLMs are sensitive to the strategy taken in few-shot learning. 
To evaluate the influence of different few-shot setups 
in terms of the number of shots and prompts
during the data generation stage on the final result, 
we explore the following setups: 
 (1) changing the number of shots from 8 to 16, the details of which are shown in Appendix \ref{fig:caption_few_shot_2}; (2) changing few-shot prompts: we pair prompt originally designed for caption with the 
 multi-genre examples. 
 

We employ GPT4 to label the data using two strategies. Then, we use the revised training data to train SimCSE-RoBERTabase models. The results are presented in Table \ref{ablation_few_shot_type}.
It can be observed that the final results on STS are not affected  by altering the prompt during the training stage. The reason for this is as follows:
STS-GPT is trained on a large amount of data points generated by LLMs, making it more resistant to prompt changes.
In contrast, the in-context learning strategy, which directly outputs results from few-shot prompting, is highly sensitive to prompt modifications.
This demonstrates the superiority of the proposed model over the in-context learning baseline.

\section{Conclusion}
\label{conclusion}
In this paper, we propose Sim-GPT, the strategy of measuring text similarity via GPT annotated data. Sim-GPT  utilizes GPT-4 to provide a substantial amount of reliable annotated data filling the gap of the lack of training signals for STS. Furthermore, we conduct experiments and analyses on 7 STS tasks: Sim-GPT not only achieves an average SOTA performances but also demonstrates consistent stability with variations in prompts. Additionally, we release the dataset consisting of 371K annotated STS examples from GPT-4 to encourage further advancements in semantic textual similarity.

\bibliography{anthology,custom}

\appendix

\section{Prompt Design}
\label{appendix_prompt}
As describe in Section \ref{data_annotation}, we utilize GPT-4 to generate a relatively large set of training data from multi-genre sources. In this section, we clearly show the prompt used in Sim-GPT, where Figure \ref{fig:caption_prompt_example} for caption sources, Figure \ref{fig:question_prompt_example} for question sources and Figure \ref{fig:multi_genre_prompt_example} for multi-genre sentence sources.

\begin{figure*}
	\begin{framed}
This task will involve reading a line from a caption and writing two sentences that relate to it. Using only this description and what you know about the world:
\begin{itemize}
		\item Write one sentence that is definitely similar to the situation or event in the line.
		\item Write one sentence that is definitely dissimilar to the situation or event in the line.
	\end{itemize}
Note that don't simply paraphrase or negate the given sentence, and we don't allow reuse the same sentence for many different prompts. \\

\underline{\textbf{Input:}} This church choir sings to the masses as they sing joyous songs from the book at a church. \\
\underline{\textbf{Output:}} 
1. The church is filled with song. \\
2. A choir singing at a baseball game. 

\underline{\textbf{Input:}} A woman with a green headscarf, blue shirt and a very big grin. \\
\underline{\textbf{Output:}} 
1. The woman is very happy. \\
2. The woman has been shot. 

\underline{\textbf{Input:}} An old man with a package poses in front of an advertisement. \\
\underline{\textbf{Output:}} 
1. A man poses in front of an ad. \\
2. A man walks by an ad. 

\underline{\textbf{Input:}} A statue at a museum that no seems to be looking at. \\
\underline{\textbf{Output:}} 
1. There is a statue that not many people seem to be interested in. \\
2. Tons of people are gathered around the statue. 

\underline{\textbf{Input:}} A land rover is being driven across a river. \\
\underline{\textbf{Output:}} 
1. A vehicle is crossing a river. \\
2. A sedan is stuck in the middle of a river. 

\underline{\textbf{Input:}} A man playing an electric guitar on stage. \\
\underline{\textbf{Output:}} 
1. A man playing guitar on stage. \\
2. A man playing banjo on the floor. 

\underline{\textbf{Input:}} One tan girl with a wool hat is running and leaning over an object, while another person in a wool hat is sitting on the ground. \\
\underline{\textbf{Output:}} 
1. A tan girl runs leans over an object. \\
2. A boy runs into a wall. 

\underline{\textbf{Input:}} Two teenage girls conversing next to lockers. \\
\underline{\textbf{Output:}} 
1. People talking next to lockers. \\
2. Girls talking next to the toilet. 
		
    \end{framed}
\caption{The designed prompt used for caption sources.}
\label{fig:caption_prompt_example}
\end{figure*}

\begin{figure*}
	\begin{framed}
This task will involve reading a line from a question and writing two questions that relate to it. The line will describe a situation or event. Using only this description and what you know about the world:
\begin{itemize}
		\item Write one question is definitely similar to the situation or event in the line.
		\item Write one question is definitely dissimilar to the situation or event in the line.
	\end{itemize}
Note that don't simply paraphrase or negate the given sentence, and we don't allow reuse the same sentence for many different prompts. \\

\underline{\textbf{Input:}} How did Michel Foucault feel about surveillance society? \\
\underline{\textbf{Output:}} 
1. What is Michel Foucaults opinion on surveillance society? \\
2. What are the main ingredients in a traditional apple pie?

\underline{\textbf{Input:}} At what time should I drink green tea to be fit? \\
\underline{\textbf{Output:}} 
1. What is the best time to drink green tea for fitness? \\
2. Does Lipton green tea Assist in weight loss? 

\underline{\textbf{Input:}} Is Quora becoming the new Facebook? \\
\underline{\textbf{Output:}} 
1. Is Quora becoming a new Facebook? \\
2. Does Quora have nothing in common with Facebook?

\underline{\textbf{Input:}} Is 12 Mbps a good internet speed for streaming and gaming? \\
\underline{\textbf{Output:}} 
1. Would I be able to stream and game effectively with a 12 Mbps internet connection? \\
2. Is 6 Mbps download and 2 Mbps upload speed good for a 4G LTE connection?

\underline{\textbf{Input:}} How can I increase my website page rank? \\
\underline{\textbf{Output:}} 
1. How do I increase page rank of website? \\
2. What techniques can lower my website's page rank? 

\underline{\textbf{Input:}} 2 states movie download? \\
\underline{\textbf{Output:}} 
1. Where can I download the movie "2 States"? \\
2. Who controls the U.S president?

\underline{\textbf{Input:}} What direction are you going when travelling from Malaga to Gibraltar? \\
\underline{\textbf{Output:}} 
1. What direction are you travelling from malaga to gibralter? \\
2. What direction are you going when travelling from Gibraltar to Malaga?

\underline{\textbf{Input:}} Do animal cells have cell walls? \\
\underline{\textbf{Output:}} 
1. Can you find cell walls in animal cells? \\
2. Is anatomy destiny? 
		
    \end{framed}
\caption{The designed prompt used for questions sources.}
\label{fig:question_prompt_example}
\end{figure*}

\begin{figure*}
	\begin{framed}
This task will involve reading a line from a non-fiction article and writing two sentences that relate to it. The line will describe a situation or event. Using only this description and what you know about the world:
\begin{itemize}
		\item Write one sentence that is definitely similar to the situation or event in the line.
		\item Write one sentence that is definitely dissimilar to the situation or event in the line.
	\end{itemize}
Note that don't simply paraphrase or negate the given sentence, and we don't allow reuse the same sentence for many different prompts. \\

\underline{\textbf{Input:}} Your contribution helped make it possible for us to provide our students with a quality education. \\
\underline{\textbf{Output:}} 
1. Your support helped us provide students with a quality education. \\
2. Your contributions were of no help with our students' education.

\underline{\textbf{Input:}} They were promptly executed. \\
\underline{\textbf{Output:}} 
1. They were executed. \\
2. They were set free, uninjured.

\underline{\textbf{Input:}} The simplest is for one or more of the members to simply donate one million dollars to the IGGS Scholarship Fund. \\
\underline{\textbf{Output:}} 
1. The simplest is for one million dollars to be donated by one of the members to the IGGS Scholarship Fund. \\
2. The most complicated is to arrange a donation of one million dollars to the IGGS Scholarship Fund.

\underline{\textbf{Input:}} In the felled seams of shirts and jeans, for example, the visible stitches might be of a color designed to decorate the garment. \\
\underline{\textbf{Output:}} 
1. The felled seams of shirts are visible. \\
2. The felled seams of shirts are not visible.

\underline{\textbf{Input:}} In this competition for legitimacy, secular regimes had no alternative to offer. \\
\underline{\textbf{Output:}} 
1. Secular regimes had little to off in way of legitimacy. \\
2. Secular regimes had many options to offer.

\underline{\textbf{Input:}} In this period, a number of key technological changes  sewing machines that made many more stitches a minute, long knives instead of shears for cutting, and pressing machines. \\
\underline{\textbf{Output:}} 
1. In this period, various key innovative changes, sewing machines that made numerous more lines a moment, long blades rather than shears for cutting, and squeezing machines. \\
2. Industry tools and machineries have stayed the same until today.

\underline{\textbf{Input:}} At the end of the second year, children begin to label their own and others' internal states with words, such as want, happy, mad, think, and pretend. \\
\underline{\textbf{Output:}} 
1. Usually by the end of the second year, children will begin to label their own and other's internal states with words. \\
2. Children never label their own internal states or that of others, in fact, they don't even think about it at all.

\underline{\textbf{Input:}} Eleven, the first guy, he's heading towards Washington. \\
\underline{\textbf{Output:}} 
1. Eleven is DC bound. \\
2. The first one was bound for Boston. 
		
    \end{framed}
\caption{The designed prompt used for multi-genre long sentences sources.}
\label{fig:multi_genre_prompt_example}
\end{figure*}

\begin{figure*}
\footnotesize
	\begin{framed}
This task will involve reading a line from a caption and writing two sentences that relate to it. Using only this description and what you know about the world:
\begin{itemize}
		\item Write one sentence that is definitely similar to the situation or event in the line.
		\item Write one sentence that is definitely dissimilar to the situation or event in the line.
	\end{itemize}
Note that don't simply paraphrase or negate the given sentence, and we don't allow reuse the same sentence for many different prompts. \\

\underline{\textbf{Input:}} A biker races. \\
\underline{\textbf{Output:}} 
1. A person is riding a bike. \\
2. The car is yellow. 

\underline{\textbf{Input:}} A woman with a green headscarf, blue shirt and a very big grin. \\
\underline{\textbf{Output:}} 
1. The woman is very happy. \\
2. The woman has been shot. 

\underline{\textbf{Input:}} A softball player throws the ball to her teammate. \\
\underline{\textbf{Output:}} 
1. Two people are playing softball. \\
2. Two softball players are sitting on a bench.

\underline{\textbf{Input:}} Island native fishermen reeling in their nets after a long day's work. \\
\underline{\textbf{Output:}} 
1. The men are finishing their day of work. \\
2. The men did not go to work today but instead played bridge. 

\underline{\textbf{Input:}} Everyone on the street in the city seem to be busy doing their own thing, \\
\underline{\textbf{Output:}} 
1. People are conducting business in the city. \\
2. The people are watching what is happening.

\underline{\textbf{Input:}} People are on an escalator waiting to get to their destination while looking outside of the glass that makes up the wall. \\
\underline{\textbf{Output:}} 
1. People are riding on the escalator. \\
2. People are taking the elevator. 

\underline{\textbf{Input:}} These are young adults who seem to be working together to protect the plants surrounding the white pole. \\
\underline{\textbf{Output:}} 
1. The adults are young. \\
2. The adults are old. 

\underline{\textbf{Input:}} Children in yellow sports uniforms climbing a tower. \\
\underline{\textbf{Output:}} 
1. Children in uniforms climb a tower. \\
2. The kids crawl in sand. 

\underline{\textbf{Input:}} This church choir sings to the masses as they sing joyous songs from the book at a church. \\
\underline{\textbf{Output:}} 
1. The church is filled with song. \\
2. A choir singing at a baseball game. 

\underline{\textbf{Input:}} A woman with a green headscarf, blue shirt and a very big grin. \\
\underline{\textbf{Output:}} 
1. The woman is very happy. \\
2. The woman has been shot. 

\underline{\textbf{Input:}} An old man with a package poses in front of an advertisement. \\
\underline{\textbf{Output:}} 
1. A man poses in front of an ad. \\
2. A man walks by an ad. 

\underline{\textbf{Input:}} A statue at a museum that no seems to be looking at. \\
\underline{\textbf{Output:}} 
1. There is a statue that not many people seem to be interested in. \\
2. Tons of people are gathered around the statue. 

\underline{\textbf{Input:}} A land rover is being driven across a river. \\
\underline{\textbf{Output:}} 
1. A vehicle is crossing a river. \\
2. A sedan is stuck in the middle of a river. 

\underline{\textbf{Input:}} A man playing an electric guitar on stage. \\
\underline{\textbf{Output:}} 
1. A man playing guitar on stage. \\
2. A man playing banjo on the floor. 

\underline{\textbf{Input:}} One tan girl with a wool hat is running and leaning over an object, while another person in a wool hat is sitting on the ground. \\
\underline{\textbf{Output:}} 
1. A tan girl runs leans over an object. \\
2. A boy runs into a wall. 

\underline{\textbf{Input:}} Two teenage girls conversing next to lockers. \\
\underline{\textbf{Output:}} 
1. People talking next to lockers. \\
2. Girls talking next to the toilet. 
		
    \end{framed}
\caption{The designed prompt with 16-shot examples used for caption sources.}
\label{fig:caption_few_shot_2}
\end{figure*}

\begin{figure*}
	\begin{framed}
This task will involve reading a line from a caption and writing two sentences that relate to it. Using only this description and what you know about the world:
\begin{itemize}
		\item Write one sentence that is definitely similar to the situation or event in the line.
		\item Write one sentence that is definitely dissimilar to the situation or event in the line.
	\end{itemize}
Note that don't simply paraphrase or negate the given sentence, and we don't allow reuse the same sentence for many different prompts. \\

\underline{\textbf{Input:}} Your contribution helped make it possible for us to provide our students with a quality education. \\
\underline{\textbf{Output:}} 
1. Your support helped us provide students with a quality education. \\
2. Your contributions were of no help with our students' education. 

\underline{\textbf{Input:}} At the end of the second year, children begin to label their own and others' internal states with words, such as want, happy, mad, think, and pretend. \\
\underline{\textbf{Output:}} 
1. Usually by the end of the second year, children will begin to label their own and other's internal states with words. \\
2. Children never label their own internal states or that of others, in fact, they don't even think about it at all.

\underline{\textbf{Input:}} They were promptly executed. \\
\underline{\textbf{Output:}} 
1. They were executed. \\
2. They were set free, uninjured.

\underline{\textbf{Input:}} The simplest is for one or more of the members to simply donate one million dollars to the IGGS Scholarship Fund. \\
\underline{\textbf{Output:}} 
1. The simplest is for one million dollars to be donated by one of the members to the IGGS Scholarship Fund. \\
2. The most complicated is to arrange a donation of one million dollars to the IGGS Scholarship Fund.

\underline{\textbf{Input:}} In the felled seams of shirts and jeans, for example, the visible stitches might be of a color designed to decorate the garment. \\
\underline{\textbf{Output:}} 
1. The felled seams of shirts are visible. \\
2. The felled seams of shirts are not visible.

\underline{\textbf{Input:}} In this competition for legitimacy, secular regimes had no alternative to offer. \\
\underline{\textbf{Output:}} 
1. Secular regimes had little to off in way of legitimacy. \\
2. Secular regimes had many options to offer. 

\underline{\textbf{Input:}} In this period, a number of key technological changes  sewing machines that made many more stitches a minute, long knives instead of shears for cutting, and pressing machines. \\
\underline{\textbf{Output:}} 
1. In this period, various key innovative changes, sewing machines that made numerous more lines a moment, long blades rather than shears for cutting, and squeezing machines. \\
2. Industry tools and machineries have stayed the same until today.

\underline{\textbf{Input:}} Eleven, the first guy, he's heading towards Washington. \\
\underline{\textbf{Output:}} 
1. Eleven is DC bound. \\
2. The first one was bound for Boston. 
		
    \end{framed}
\caption{The designed prompt with 8-shot multi-genre examples for caption sources.}
\label{fig:caption_few_shot_3}
\end{figure*}

\begin{figure*}
	\begin{framed}
This task is the semantic textual similarity task. You will read two sentences and write a similarity score of them. The score ranges from 0 to 5, including decimal values, where 0 indicates no similarity and 5 represents the highest level of similarity. \\

\underline{\textbf{Input:}} a man with a hard hat is dancing .\\a man wearing a hard hat is dancing . \\
\underline{\textbf{Output:}} 5.0

\underline{\textbf{Input:}} a woman is playing the guitar .\\a man is playing guitar . \\
\underline{\textbf{Output:}} 2.4

\underline{\textbf{Input:}} people are playing cricket .\\men are playing cricket .\\
\underline{\textbf{Output:}} 3.2

\underline{\textbf{Input:}} a man is speaking .\\a man is spitting . \\
\underline{\textbf{Output:}} 0.636

\underline{\textbf{Input:}} a man is eating food .\\a man is eating something . \\
\underline{\textbf{Output:}} 4.2

\underline{\textbf{Input:}} the man is riding a horse .\\a woman is using a hoe . \\
\underline{\textbf{Output:}} 0

\underline{\textbf{Input:}} the boy is playing the piano .\\a band is playing on stage . \\
\underline{\textbf{Output:}} 1.333

\underline{\textbf{Input:}} a man and a woman walk through the woods .\\the man and woman are walking . \\
\underline{\textbf{Output:}} 3.0
		
    \end{framed}
\caption{The designed 8-shot prompt used for in-context learning-based method.}
\label{fig:context_learning_8}
\end{figure*}

\begin{figure*}
	\begin{framed}
This task is the semantic textual similarity task. You will read two sentences and write a similarity score of them. The score ranges from 0 to 5, including decimal values, where 0 indicates no similarity and 5 represents the highest level of similarity. \\

\underline{\textbf{Input:}} a man with a hard hat is dancing .\\a man wearing a hard hat is dancing . \\
\underline{\textbf{Output:}} 5.0

\underline{\textbf{Input:}} a woman is playing the guitar .\\a man is playing guitar . \\
\underline{\textbf{Output:}} 2.4

\underline{\textbf{Input:}} people are playing cricket .\\men are playing cricket .\\
\underline{\textbf{Output:}} 3.2

\underline{\textbf{Input:}} a man is speaking .\\a man is spitting . \\
\underline{\textbf{Output:}} 0.636

\underline{\textbf{Input:}} a man is eating food .\\a man is eating something . \\
\underline{\textbf{Output:}} 4.2

\underline{\textbf{Input:}} the man is riding a horse .\\a woman is using a hoe . \\
\underline{\textbf{Output:}} 0

\underline{\textbf{Input:}} the boy is playing the piano .\\a band is playing on stage . \\
\underline{\textbf{Output:}} 1.333

\underline{\textbf{Input:}} a man and a woman walk through the woods .\\the man and woman are walking . \\
\underline{\textbf{Output:}} 3.0

\underline{\textbf{Input:}} the lady stirred up raw eggs in the bowl .\\a woman is pouring eyes into a bowl . \\
\underline{\textbf{Output:}} 1.0

\underline{\textbf{Input:}} a cow is eating grass .\\a dog is pulling a girl down a hill . \\
\underline{\textbf{Output:}} 0.0

\underline{\textbf{Input:}} the man stirred the sauce for the chicken .\\the man is stirring oil . \\
\underline{\textbf{Output:}} 2.4

\underline{\textbf{Input:}} the lady cut the tail and body of a shrimp .\\a woman is cleaning a shrimp . \\
\underline{\textbf{Output:}} 4.5

\underline{\textbf{Input:}} a person is peeling shrimp .\\a person is preparing shrimp . \\
\underline{\textbf{Output:}} 3.6

\underline{\textbf{Input:}} the man is slicing a potato .\\a man is slicing potato . \\
\underline{\textbf{Output:}} 5.0

\underline{\textbf{Input:}} ta man is cutting up a potato .\\a man is cutting up carrots . \\
\underline{\textbf{Output:}} 2.375

\underline{\textbf{Input:}} the man is hiking in the woods .\\a man is tracking in the wood . \\
\underline{\textbf{Output:}} 3.0
		
    \end{framed}
\caption{The designed 16-shot prompt used for in-context learning-based method.}
\label{fig:context_learning_16}
\end{figure*}

\begin{figure*}
	\begin{framed}
This task is the semantic textual similarity task. You will read two sentences and write a similarity score of them. The score ranges from 0 to 5, including decimal values, where 0 indicates no similarity and 5 represents the highest level of similarity. \\

\underline{\textbf{Input:}} a man with a hard hat is dancing .\\a man wearing a hard hat is dancing . \\
\underline{\textbf{Output:}} 5.0

\underline{\textbf{Input:}} a woman is playing the guitar .\\a man is playing guitar . \\
\underline{\textbf{Output:}} 2.4

\underline{\textbf{Input:}} people are playing cricket .\\men are playing cricket .\\
\underline{\textbf{Output:}} 3.2

\underline{\textbf{Input:}} a man is speaking .\\a man is spitting . \\
\underline{\textbf{Output:}} 0.636

\underline{\textbf{Input:}} a man is eating food .\\a man is eating something . \\
\underline{\textbf{Output:}} 4.2

\underline{\textbf{Input:}} the man is riding a horse .\\a woman is using a hoe . \\
\underline{\textbf{Output:}} 0

\underline{\textbf{Input:}} the boy is playing the piano .\\a band is playing on stage . \\
\underline{\textbf{Output:}} 1.333

\underline{\textbf{Input:}} a man and a woman walk through the woods .\\the man and woman are walking . \\
\underline{\textbf{Output:}} 3.0

\underline{\textbf{Input:}} the lady stirred up raw eggs in the bowl .\\a woman is pouring eyes into a bowl . \\
\underline{\textbf{Output:}} 1.0

\underline{\textbf{Input:}} a cow is eating grass .\\a dog is pulling a girl down a hill . \\
\underline{\textbf{Output:}} 0.0

\underline{\textbf{Input:}} the man stirred the sauce for the chicken .\\the man is stirring oil . \\
\underline{\textbf{Output:}} 2.4

\underline{\textbf{Input:}} the lady cut the tail and body of a shrimp .\\a woman is cleaning a shrimp . \\
\underline{\textbf{Output:}} 4.5

\underline{\textbf{Input:}} a person is peeling shrimp .\\a person is preparing shrimp . \\
\underline{\textbf{Output:}} 3.6

\underline{\textbf{Input:}} the man is slicing a potato .\\a man is slicing potato . \\
\underline{\textbf{Output:}} 5.0

\underline{\textbf{Input:}} ta man is cutting up a potato .\\a man is cutting up carrots . \\
\underline{\textbf{Output:}} 2.375

\underline{\textbf{Input:}} the man is hiking in the woods .\\a man is tracking in the wood . \\
\underline{\textbf{Output:}} 3.0
		
    \end{framed}
\end{figure*}

\begin{figure*}
	\begin{framed}

\underline{\textbf{Input:}} a man is playing guitar .\\a man plays a guitar . \\
\underline{\textbf{Output:}} 4.857

\underline{\textbf{Input:}} the man is buttering the bread .\\the man is stirring the rice . \\
\underline{\textbf{Output:}} 0.4

\underline{\textbf{Input:}} large silver locomotive engine in a shed .\\the silver train is parked in a station . \\
\underline{\textbf{Output:}} 2.6

\underline{\textbf{Input:}} the back of a stop sign with many stickers on it .\\the back of a sign with stickers on . \\
\underline{\textbf{Output:}} 3.8

\underline{\textbf{Input:}} a doubly decker red bus driving down the road .\\a red double decker bus driving down a street . \\
\underline{\textbf{Output:}} 5

\underline{\textbf{Input:}} the black bird is sitting on the ground .\\the back of a pig under a tree with a cow in the background . \\
\underline{\textbf{Output:}} 0

\underline{\textbf{Input:}} a black and white horned cow standing in a field .\\a large black and white cow in a field . \\
\underline{\textbf{Output:}} 4

\underline{\textbf{Input:}} a red bird and four other birds sitting in the snow .\\five birds stand on the snow . \\
\underline{\textbf{Output:}} 2.8

\underline{\textbf{Input:}} domestic cat looking out window .\\a white cat looking out of a window . \\
\underline{\textbf{Output:}} 3.6

\underline{\textbf{Input:}} two kids are playing a game of foosball .\\the kids are playing a game with each other . \\
\underline{\textbf{Output:}} 3.4

\underline{\textbf{Input:}} the man is in a deserted field .\\the man is outside in the field . \\
\underline{\textbf{Output:}} 4.0

\underline{\textbf{Input:}} a musician is smearing jam on his white guitar at a concert .\\trombonist playing the her instrument in a band for a parade . \\
\underline{\textbf{Output:}} 0.4

\underline{\textbf{Input:}} a pair of young boys in t-shirts are hiding in the woods with one looking aghast .\\two smiling little girls playing in a fountain with other people . \\
\underline{\textbf{Output:}} 0.0

\underline{\textbf{Input:}} one football player tries to tackle a player on the opposing team .\\a football player attempts a tackle . \\
\underline{\textbf{Output:}} 4.6

\underline{\textbf{Input:}} there are people out on the street .\\people are out on the street . \\
\underline{\textbf{Output:}} 5.0

\underline{\textbf{Input:}} all we know is this : distant objects are receding from us at a rate proportional to their distance from us .\\the expansion of space means that objects in cosmological distances are receding away from each other . \\
\underline{\textbf{Output:}} 3.4
		
    \end{framed}
\caption{The designed 32-shot prompt used for in-context learning-based method.}
\label{fig:context_learning_32}
\end{figure*}

\end{document}